\documentclass[accepted]{uai2023}
\usepackage[american]{babel}

\usepackage{natbib} 
    \renewcommand{\bibsection}{\subsubsection*{References}}

\usepackage{hyperref}
\usepackage{xurl}
\usepackage{booktabs}       
\usepackage{amsfonts}       
\usepackage{nicefrac}       
\usepackage{microtype}      
\usepackage{xcolor}         
\usepackage{multirow}
\usepackage{multicol}
\usepackage{subcaption}
\usepackage{siunitx} 
\usepackage{amsmath}
\usepackage{amssymb}
\usepackage{mathtools}
\usepackage{amsthm}
\usepackage{caption}
\usepackage{enumitem}
\usepackage{pdfpages}

\MakeRobust{\ref}

\makeatletter
\newcommand{\labeltext}[2]{%
  \@bsphack \csname phantomsection\endcsname 
  \def\@currentlabel{#1}{\label{#2}}%
  \@esphack }
\makeatother

\title{SPDF: Sparse Pre-training and Dense Fine-tuning for Large Language Models}

\author[1]{\href{mailto:vithu@cerebras.net}{Vithursan Thangarasa}{}}
\author[1]{Abhay~Gupta}
\author[1]{William~Marshall}
\author[*]{Tianda~Li}
\author[1]{Kevin~Leong} 
\author[*]{\\Dennis DeCoste}
\author[1]{Sean Lie}
\author[1]{Shreyas Saxena}
\affil[1]{%
    Cerebras Systems Inc.\\
    Sunnyvale, California, USA }
\begin{document}

\maketitle

\begin{abstract}
    The pre-training and fine-tuning paradigm has contributed to a number of
    breakthroughs in Natural Language Processing (NLP). Instead of directly
    training on a downstream task, language models are first pre-trained on
    large datasets with cross-domain knowledge (e.g., Pile, MassiveText, etc.)
    and then fine-tuned on task-specific data (e.g., natural language
    generation, text summarization, etc.). Scaling the model and dataset size
    has helped improve the performance of LLMs, but unfortunately, this also
    lead to highly prohibitive computational costs. Pre-training LLMs often
    require orders of magnitude more FLOPs than fine-tuning and the model
    capacity often remains the same between the two phases. To achieve training
    efficiency~w.r.t training FLOPs, we propose to decouple the model capacity
    between the two phases and introduce Sparse Pre-training and Dense
    Fine-tuning (SPDF). In this work, we show the benefits of using unstructured
    weight sparsity to train only a subset of weights during pre-training
    (Sparse Pre-training) and then recover the representational capacity by
    allowing the zeroed weights to learn (Dense Fine-tuning). We demonstrate
    that we can induce up to 75\% sparsity into a 1.3B parameter GPT-3 XL model
    resulting in a 2.5x reduction in pre-training FLOPs, without a significant
    loss in accuracy on the downstream tasks relative to the dense baseline. By
    rigorously evaluating multiple downstream tasks, we also establish a
    relationship between sparsity, task complexity and dataset size. Our work
    presents a promising direction to train large GPT models at a fraction of
    the training FLOPs using weight sparsity, while retaining the benefits of
    pre-trained textual representations for downstream tasks.
    \footnotetext[0]{\textsuperscript{*}Work done while at Cerebras Systems.}
\end{abstract}

\section{Introduction}
Large language models (LLMs) have contributed to significant advances in natural
language understanding (NLU) and natural language generation (NLG) due to the
introduction of pre-training methods~\citep{Devlin2019BERTPO,
Radford2018ImprovingLU} on massive unannotated datasets (e.g.,
Pile~\citep{gao2020pile}, MassiveText~\citep{rae2021gopher}, etc.). While
scaling the model and dataset size has improved the quality of
LLMs~\citep{wei2022emergent}, it has also substantially increased the
computational cost of pre-training. For instance, GPT-3
175B~\citep{brown2020gpt3} is estimated to cost millions of dollars to
train~\citep{li_2022}. Various techniques have been proposed to reduce the
computational cost of training LLMs, including sparse
attention~\citep{dao2022flashattention, jaszczur2021sparse}, improved
optimization techniques~\citep{tang20221bitadam} and sequence-level curriculum
learning~\citep{li2022the}. While these methods can help reduce computation
time, weight sparsity is one promising technique orthogonal to the above
methods. Here, a subset of model parameters are set to zero, reducing the FLOPs
required during training. 

Despite recent advances in sparse training~\citep{hoefler2022sparsity}, it has
yet to be widely adopted by practitioners. First, it is difficult and expensive
to find the optimal sparsity pattern~\citep{frankle2018lottery, ma2022effective}
that can maintain the same level of accuracy as dense models. Second,
unstructured sparsity can be difficult to accelerate on hardware architectures
optimized for dense computation~\citep{sara2020lottery}. In this work, we show
how we can leverage weight sparsity to reduce training FLOPs, and then recover
the lost representational capacity by shifting to dense weight matrices when
fine-tuning on downstream tasks. In addition, while specialized software kernels
have been developed to achieve inference acceleration with unstructured
sparsity~\citep{gale2020sparse, neural_magic_2021, elsen2019sparse,
Ashby2019ExploitingUS, Wang2021SparseDNNFS}, recent work has shown that we can
realize the gains of unstructured weight sparsity on specialized hardware (e.g.,
Cerebras CS-2~\citep{lie_2022, lie_2021}) when training LLMs. For
example,~\citet{lie_2021} shows the measured speedup for a matrix multiplication
kernel~w.r.t to the sparsity level on a single GPT-3 layer (see Appendix C for
more details). Therefore, as unstructured sparse training techniques continue to
become co-designed with the hardware, we can expect the FLOP reduction to
translate into performance and wall-clock speedups.

Prior work on sparsifying LLMs focus on reducing
training~\citep{chen2022pixelated, dao2022monarch} or inference
FLOPs~\citep{chen2020lth}, while matching standard dense training.
\citet{chen2022pixelated} and~\citet{dao2022monarch} replace dense matrices with
butterfly-based structured sparse weight matrices to reduce a model's size and
accelerate pre-training on block-oriented hardware (e.g.,
GPUs~\citep{krashinsky_2020}, TPUs~\citep{xin2020sparsetpu}). Training with
structured sparsity requires maintaining a regular sparse structure, which can
reduce expressivity at higher sparsity levels. This is a well-known constraint
observed when imposing structured sparsity in dense weight
matrices~\citep{zhou2021learning, jiang2022exposing}. The recent innovations in
hardware architectures aim to facilitate the widespread use and adoption of
unstructured weight sparsity, enabling the ability to achieve higher compression
ratios while attaining practical speedups~w.r.t wall-clock time. Our work
focuses on pre-training with unstructured weight sparsity to reduce the FLOPs
for training language models.

In the recent NLP literature, it is common to first pre-train, then fine-tune a
language model. Fine-tuning pre-trained LLMs on downstream tasks leads to
significantly better accuracy than the zero or few-shot
settings~\citep{alt-etal-2019-fine, ouyang2022training}. The pre-training phase
takes significantly longer compared to fine-tuning on a much smaller dataset to
learn the domain-specific task. In the standard setup, the model size and
capacity is generally kept the same between the two phases. We propose to break
this assumption and show the benefits of modifying the model capacity between
pre-training and fine-tuning with weight sparsity. First, we pre-train a sparse
GPT model to reduce computational training FLOPs. Then, during the fine-tuning
phase, we densify the GPT model, allowing the zeroed weights to learn and
increase the modelling capacity to more accurately learn the downstream task.

While previous work has explored sparse-to-dense training to mitigate the
difficulties of sparse-to-sparse training~\citep{dao2022monarch} and improve the
accuracy of dense models~\citep{han2017dsd}, we perform fully sparse
pre-training and only transition to dense weight matrices during fine-tuning. We
refer to this framework as Sparse Pre-training and Dense Fine-tuning (SPDF) and
demonstrate the ability of the sparse pre-trained model to transfer effectively
to different downstream tasks (e.g., natural language generation and text
summarization). The main contributions of our work are:

\begin{enumerate}
    \item We propose Sparse Pre-training and Dense Fine-tuning (SPDF) as a new
    framework to reduce the FLOPs required during the pre-training phase, while
    maintaining accuracy on downstream tasks.

    \item We demonstrate that we can train GPT-3 XL, at 75\% sparsity, reducing
    the overall training FLOPS by 2.5x, while retaining the benefits of
    pre-trained textual representations in LLMs across a majority of tasks and
    evaluation metrics.

    
    \item We establish a correlation between the optimal sparsity level during
    pre-training and the fine-tuning dataset size and task difficulty.
\end{enumerate}

\section{Methodology}
\label{sec:method}

This section presents our method to reduce pre-training FLOPs using unstructured
weight sparsity. We first explain our intuition and hypotheses, followed by our
methodology for the SPDF framework.

\subsection{Intuition and Hypotheses}
\label{sec:hypotheses}
Prior works have shown that overparameterization of neural networks improves
optimization and generalizability~\citep{Soltanolkotabi2019, neyshabur2018the,
pmlr-v97-allen-zhu19a}, but leads to an increase in compute
cost~\citep{brown2020gpt3}. Recent work on the Lottery Ticket
Hypothesis~\cite{frankle2018lottery} demonstrates that overparameterized dense
networks contain sparse subnetworks which can be trained to the same accuracy as
their dense counterparts, as long as one initializes the training with a good
sparsity mask (``lottery ticket''). However, the process of searching for highly
quality sparse subnetworks is computationally
expensive~\citep{frankle2018lottery, ma2022effective}. Existing sparse training
methods~\citep{evci2020rigl, mocanu2018, jayakumar2020top} aim to discover the
winning lottery ticket (i.e., optimal sparsity mask) in a single training run,
but often fall short of the dense model's accuracy. 

In our framework, we mitigate the loss in representational power due to
difficulties in sparse optimization~\citep{Evci2019TheDO}, by transitioning to
fully dense weight matrices during the fine-tuning phase. Even though we perform
dense fine-tuning, the computational costs associated with fine-tuning are
significantly lower than the cost of pre-training LLMs. Therefore, our method
targets the phase which dominates the training FLOPs (i.e., pre-training). Based
on recent theoretical findings and empirical studies on overparameterization and
sparse neural networks, we lay out a set of hypotheses which we aim to study in
our work through extensive experimental evaluation:

\paragraph*{\normalfont\textit{Hypothesis 1\labeltext{1}{hyp:one}: High degrees
of weight sparsity can be used during the pre-training phase of LLMs while
preserving the downstream accuracy with dense fine-tuning.}} \mbox{}

Inducing sparsity during pre-training may cause a loss in representational power
due to difficulties in sparse optimization and inability to discover optimal
sparsity masks~\citep{Evci2019TheDO}. To mitigate these challenges, we aim to
increase the representational power by allowing the zeroed weights to grow
during fine-tuning (i.e., dense fine-tuning). 

Additionally, note the full capacity of the pre-trained model is often not
required to generalize on the simpler downstream task, when using sparsity
during pre-training~\citep{Ding2022DeltaTA}.
\citet{aghajanyan-etal-2021-intrinsic} investigate this phenomenon from a
different angle and show pre-trained language models can learn a large set of
NLP tasks with only a few parameters.
This indicates that the full parameterization of the model is not needed to
generalize well across downstream fine-tuning tasks. Hence, we can exploit
weight sparsity during pre-training while retaining important textual
representations despite the model's lower representational capacity.

\paragraph*{\normalfont \textit{Hypothesis 2\labeltext{2}{hyp:two}: The
performance of the sparse pre-trained model is correlated with the dataset size
and degree of difficulty in the downstream task.}} \label{hyp:two} \mbox{}

\citet{liu2023sparsity} evaluate sparse networks on a diverse set of tasks with
varying degrees of difficulty and show a strong correlation between a model's
ability to be sparsified and the task difficulty. Hence, we hypothesize that
models trained on complex tasks with high sparsity levels can suffer more from
sparse training and experience a greater drop in performance compared to simpler
tasks. We also note that small fine-tuning datasets may trigger
over-fitting~\citep{Li2021ImprovedRA}. Therefore, we hypothesize that larger
datasets can allow the sparse model to improve its generalization error on the
task, and recover from training with high sparsity.

\paragraph*{\normalfont\textit{Hypothesis 3\labeltext{3}{hyp:three}: As we
 increase the size of the language model, larger models become more amenable to
 higher levels of sparsity during pre-training.}} \label{hyp:three} \mbox{}

Existing work~\citep{liu2022the} has shown that the quality of a network trained
with random static sparsity (even at high sparsity levels) improves quickly to
match its dense counterpart as the network grows wider and deeper. Also, larger
models tend to have a smaller intrinsic
dimension~\citep{aghajanyan-etal-2021-intrinsic}, which suggests that all
parameters are not required to represent the average NLP task. Therefore, we
expect the gap in downstream performance between the sparse pre-trained model
and its dense counterpart to grow smaller as the size of the model increases.  

\subsection{Sparse Pre-training and Dense Fine-tuning}
Our training procedure consists of two phases. The first phase involves
pre-training a sparse language model on a large corpus of text in an
unsupervised manner. Here, we induce unstructured weight sparsity into the
neural network to reduce the pre-training FLOPs. This is followed by a dense
fine-tuning stage, where we expand the representational capacity of the model by
allowing zeroed weights to learn, and adapt to a discriminative task with
labeled data.

\paragraph*{Unsupervised Dense Pre-training} While our proposed framework is
agnostic to the training objective, we focus on autoregressive language modeling
as our motivating use case. In an autoregressive language model, the sequence
generation process is modeled as a Markov chain, where the token to be predicted
depends on all the previous tokens~\citep{bengio2003pretrain}. Hence, the
standard approach is to learn the probability distribution over sequences of
tokens from an unsupervised pre-training corpus. Given an unsupervised
pre-training corpus of tokens~$\mathcal{U} = \{u_1,
u_2,\ldots,u_{|\mathcal{U}|}\}$,  where $|\mathcal{U}|$ is the total number of
tokens. We aim to maximize the likelihood using the language modeling objective
formulated as follows,
\begin{equation}
    \label{eq:gpt}
    \mathcal{L}(\mathcal{U}) = \sum_{i=1}^{|\mathcal{U}|}\log(p(u_i | u_{i-k},\ldots,u_{i-1}, \theta)),\notag
\end{equation}

where $k$ is the size of the context window, and the conditional probability $p$
is modeled using a neural network with parameters $\theta \in \mathbb{R}^{N}$.
The parameters of the $l^{th}$ layer $\in L$ total layers are denoted as
$\theta_l$, along with the total number of parameters represented as $N_l$. We
note that the network parameters $\theta$ are considered to be dense.

\paragraph*{Unsupervised Sparse Pre-training} To induce sparsity into the
$l^{th}$ layer, we drop $s_l \in (0, 1)$ of its connections, where $s_l$ to
refer to the sparsity of layer $l$. This results in a total of $(1-s_l)N_l$
parameters. Finally, the overall sparsity of a sparse subnetwork is defined as
the ratio of zeroes to the total number of parameters in the original dense
network, i.e., $S = \frac{\sum_l^L s_lN_l}{N}$. In our sparse training setup, we
apply a binary sparsity mask $m \in \{0,1\}^{|\theta|}$ on the initial
parameters $\theta^0$, such that its initialization is $m \odot \theta^{0}$.
Here, the values 0 and 1 in the mask denote inactive (i.e., zero) and active
(i.e., non-zero) weights, respectively. As a result, the sparse language model
minimizes the following objective instead,
\begin{equation}
    \label{eq:gpt}
    \mathcal{L}(\mathcal{U}) = \sum_{i=1}^{|\mathcal{U}|}\log(p(u_i | u_{i-k},\ldots,u_{i-1}, m \odot \theta)).
\end{equation}

In our work, we focus solely on static sparsity (i.e., $m$ remains fixed
throughout training) and the weights are randomly pruned at initialization.
Specifically, we remove weights in each layer $l \in L$ randomly to the target
sparsity $s_l$. Although several works have explored generating different
layer-wise sparsity ratios at initialization (e.g.,
Erd\"os-R\'enyi-Kernel~\cite{evci2020rigl}, Ideal Gas
Quota~\citep{chen2022sparsity}, SNIP~\citep{lee2018snip},
GraSP~\citep{Wang2020Picking}, SynFlow~\citep{tanaka2020synflow}, etc.), we
focus on the simplest setup, which is uniform sparsity~\citep{gale2019state}. In
uniform sparsity, each sparsified layer is pruned to the same target sparsity
level. 

For the language model, we use GPT~\citep{radfordGPT2, brown2020gpt3} in our
experiments, which is a variant of the
Transformer~\citep{vaswani2017transformers}. We train the network with objective
shown in Eq.~\ref{eq:gpt} and AdamW~\citep{loshchilov2017decoupled} optimizer on
an unsupervised pre-training dataset for a total of $j$ iterations, arriving at
parameters $\theta^j$. Then, we adapt (i.e., fine-tune) the final pre-trained
autoregressive language model $p_{m \odot \theta}$ to the supervised target
task.

\begin{figure}[!t]
    \centering
    \includegraphics[keepaspectratio=true, width=\linewidth]{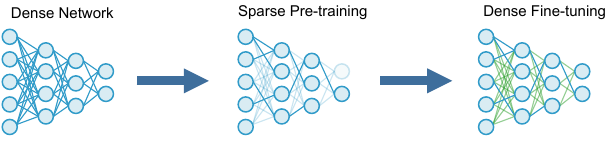}
    \caption{Sparse Pre-training and Dense Fine-tuning (SPDF) framework. In this
    framework, we sparsify a dense network and perform sparse pre-training
    followed by dense fine-tuning (green connections indicate newly activated
    weights). We use SPDF to pre-train large GPT models at a fraction of the
    training FLOPs using weight sparsity, and still retain the benefits on
    downstream tasks with dense fine-tuning.  }
    \label{fig:spdf}
    \vspace{-0.1in}
\end{figure}

\paragraph*{Dense Fine-tuning} 
Following~\citet{hu2022lora} and~\citet{Li2021PrefixTuningOC}, each downstream
fine-tuning task is represented by a training dataset consisting of
context-target pairs defined as: $\mathcal{Z} = \{(x_1 , y_1),(x_2 ,
y_2),\ldots,(x_{|x|}, y_{|y|})\}$, where both $x$ and $y$ are sequences of
tokens. For example, in structured data-to-text (e.g.,
E2E~\citep{novikona2017e2e}), $x$ corresponds to a linearized data table and $y$
a textual description; in text summarization (e.g., Curation
Corpus~\citep{curationcorpusbase:2020}), $x$ is the content of an article and
$y$ is its summary. 

We initialize the start of dense fine-tuning to the final pre-trained parameters
$\theta^j$ and during fine-tuning are updated to $\theta^j + \Delta\theta$. For
each downstream task,  we learn a different set of parameters with the
task-specific parameter increment $\Delta\theta$ whose dimension
$|\Delta\theta|$ equals $|\theta|$. Other works have explored more parameter
efficient approaches to reduce the size of the task-specific parameters for the
purpose of deploying fine-tuned models~\citep{ben-zaken-etal-2022-bitfit,
pmlr-v97-houlsby19a,hu2022lora}.  However, in our approach, we focus on reducing
the pre-training FLOPs with unstructured weight sparsity and perform dense
fine-tuning to mitigate the challenges of sparse optimization by increasing
representational power of the network. In the dense fine-tuning phase, we
essentially remove the sparsity mask $m$ to allow the inactive weights to grow.
More specifically, we increase the representational capacity in $\theta^j$ by
reviving all $\sum_{l}^{L}s_l {\cdot} N_l$ inactive weights, where all newly
activated weights are initialized to 0. We evaluated other initializations like
scaled normal distribution, but this did not lead to better results. Finally,
the network is updated in a dense manner with the objective shown below,
\begin{equation}
    \mathcal{L}(\mathcal{Z}) = \sum_{(x,y)\in\mathcal{Z}}\sum_{t=1}^{|\boldsymbol{y}|}\log(p(y_t | (x_{1},\ldots,x_{t-1}), \theta^j + \Delta\theta)).\notag
\end{equation}

The generic Sparse Pre-training and Dense Fine-Tuning (SPDF) framework,
illustrated in Figure~\ref{fig:spdf}, consists of the following three steps:
\begin{enumerate}
    \item Sparsify a given dense network to some target sparsity level, $s_l$,
    at each sparsifiable layer.
    \item \textit{Pre-train} the sparse model following the same training
    schedule as the original dense model.
    \item \textit{Fine-tune} the pre-trained sparse network on a given
    downstream task in a dense manner by allowing the zeroed weights to learn.
\end{enumerate}

\section{Experimental Setup and Results}
\label{sec:results}
First, we provide details on our pre-training settings for GPT-2 Small (125M)
and GPT-3 XL (1.3B), as well as our setups for the downstream fine-tuning tasks.
Then, we compare sparse pre-training and sparse fine-tuning against sparse
pre-training and dense fine-tuning to highlight the benefits of fine-tuning in a
dense manner. Next, we validate our hypotheses (refer to
Section~\ref{sec:hypotheses}) by evaluating SPDF across several tasks in natural
language generation and text summarization. Following this, we compare the
parameter subspaces between the pre-trained and fine-tuned models. Last, we
present the advantages in training efficiency w.r.t total training FLOPs when
using SPDF versus standard dense pre-training and dense fine-tuning. 

All GPT models are pre-trained and fine-tuned using the Cerebras CS-2, taking
advantage of its ability to accelerate training with unstructured sparsity. At
present, the specialized kernels of Cerebras CS-2 are designed to facilitate
training with static unstructured sparsity. Consequently, the results presented
in this section do not include the utilization of dynamic sparse training
methods (e.g., SET~\citep{mocanu2018}, RigL~\citep{evci2020rigl}, etc). In
Appendix C, we emphasize the possible advantages achieved through unstructured
weight sparsity on the Cerebras CS-2. We provide measured speedup results
compared to theoretical speedup across different sparsity levels for a GPT-3
layer's 12k~$\times$ 12k matrix multiplication (MatMul)~\citep{lie_2022}.

\begin{figure*}[!t]
    \centering
    \includegraphics[keepaspectratio=true, width=0.82\linewidth]{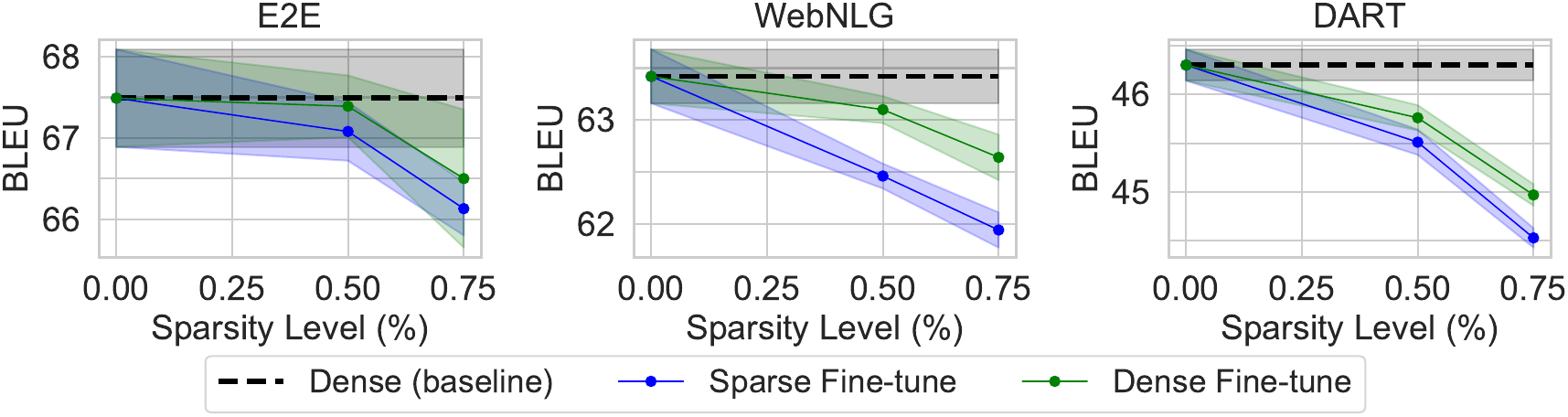}
    \caption{ Comparison of sparse-to-dense vs sparse-to-sparse pre-training and
    fine-tuning with GPT-2 Small on E2E, WebNLG and DART. Across tasks dense
    fine-tuning noticeably outperforms sparse fine-tuning, especially at 75\%
    sparsity. }
    \label{fig:sparseft}
    \vspace{-0.1in}
\end{figure*}

\paragraph{Flop Optimal Pre-training via Chinchilla Scaling Law}
It was previously conventional in the literature to train all large language
models (e.g., GPT-3~\citep{brown2020gpt3}, Gopher~\citep{rae2021gopher},
Jurassic~\citep{J1WhitePaper}, etc.) on approximately 300B tokens of data. More
recently, Chinchilla~\citep{hoffmann2022an} shows how parameters and data should
be scaled equally as compute budget increases, which leads to significant gains
in FLOP efficiency. In our pre-training setup, we follow Chinchilla's scaling
law which suggests that we need approximately 20 tokens per parameter. Thus, for
GPT-2 Small, a model with 125M parameters needs to be pre-trained on 2.5B
tokens. Then, for GPT-3 XL, a model which has 1.3B parameters, needs to be
pre-trained on 26B tokens. Unless stated otherwise, we pre-train our sparse GPT
models from scratch on the Pile dataset~\citep{gao2020pile} across sparsity
levels $S \in \{50\%, 75\%\}$.

\paragraph{Fine-tuning on Downstream Tasks}  We studied dense fine-tuning on
several downstream tasks in natural language generation and text summarization.
We follow~\citet{hu2022lora} in using the three standard natural language
generation benchmark datasets (i.e., E2E~\citep{novikona2017e2e},
WebNLG~\citep{gardent2017webnlg} and DART~\citep{nan2021dart}). In addition, we
fine-tune on Curation Corpus~\citep{curationcorpusbase:2020} according to the
details described in~\citep{rae2021gopher}. We fine-tune all parameters of the
pre-trained GPT models and evaluate the final fine-tuning performance using the
official evaluation scripts. More details on the hyperparameters can be found in
Appendix A.

\subsection{Details on the Fine-tuning Datasets}

Our work uses four fine-tuning datasets to investigate the efficacy of our SPDF
framework. These datasets were chosen for studying the effect of sparse
pre-training on different sizes and types of data, along with the varying degree
of difficulty in the tasks.

\textbf{End-2-End (E2E) NLG challenge} dataset contains approximately 45k
training, 4.6k validation, and 4.6k test examples with 8 distinct fields from
the restaurant domain. The goal of the task is to generate natural language
descriptions in the restaurant domain from meaning representations. We use the
official evaluation script, which reports BLEU~\citep{kishore2002bleu},
NIST~\citep{belz2006comparing}, METEOR~\citep{alon2007meteor},
ROUGE-L~\citep{lin2004rouge}, and CIDEr~\citep{ramakrishna2015cider}.

\textbf{WebNLG} dataset consists of 18k training, 2.2k validation, and 2.4k test
examples, where the input is a sequence of (subject, property, object) triples.
In the training and validation splits, the input describes entities from 9
distinct DBpedia categories. The test set contains 15 different domains where 10
are present only in the training data. Here, the test data is split into two
parts, where categories seen in the train set are in the first half, while the
second half consists of 5 unseen categories. We use the official evaluation
script, which reports BLEU, METEOR and TER~\citep{snover2006study}. The WebNLG
dataset is the smallest of the three NLG tasks we evaluate on.

\textbf{DART} is an open domain DAta-Record-to-Text (i.e., table-to-text)
dataset, with a similar input format to WebNLG. It consists of 62.6k training,
6.9k validation, and 12.5k test examples from several sources:
WikiSQL~\citep{zhong2018seqsql},
WikiTableQuestions~\citep{pasupat2015compositional}, Cleaned
E2E\footnote{\url{https://github.com/tuetschek/e2e-cleaning}}, and WebNLG
2017\footnote{\url{https://gitlab.com/shimorina/webnlg-dataset/-/tree/master/webnlg_challenge_2017}}
and applies some manual or automated conversion. We use the official evaluation
script and report BLEU, METEOR and TER. The DART dataset is considered to be the
most challenging NLG task out of the three we evaluate.

\textbf{Curation Corpus} is a recently introduced dataset comprised of 40,000
bespoke text summaries of finance articles for the task of text summarization.
We follow the instructions in the Curation Corpus GitHub
repository\footnote{\url{https://github.com/CurationCorp/curation-corpus}} to
download approximately 40k article summary pairs. After filtering examples where
either the article or the summary are empty, we are left with 39,911 examples.
Following~\citet{marfurt2021sentence}, we split them into train/validation/test
sets as 80/10/10 to arrive at split sizes of 31,929/3,991/3,991.

\subsection{Sparse Fine-tuning vs Dense Fine-tuning} In this section, we first
empirically establish the need for dense fine-tuning to help mitigate the
difficulties of sparse-to-sparse training (i.e., sparse pre-training followed by
sparse fine-tuning). In Figure~\ref{fig:sparseft}, we compare dense fine-tuning
against sparse fine-tuning on GPT-2 Small and show that across all three NLG
tasks (i.e., E2E, WebNLG and DART), dense fine-tuning helps reduce the drop in
BLEU score relative to the respective dense baselines. For example, the 75\%
sparse GPT-2 Small model on WebNLG observes a delta of -1.48 and -0.78 in the
BLEU scores, when sparse fine-tuning and dense fine-tuning, respectively. This
suggests that fully sparse end-to-end pre-training and fine-tuning can prevent
the model from generalizing well on downstream tasks. However, we can mitigate
the difficulties of poor generalizability due to sparse-only training by
transitioning from sparse to dense matrices during the fine-tuning phase.
Although dense fine-tuning consumes more FLOPs compared to sparse fine-tuning,
the overall fine-tuning FLOPs relative to pre-training, still remains
insignificant  (discussed further in Section~\ref{sec:spdf_train_eff}). 


\begin{table}[!t]
    \caption{Downstream accuracy of GPT-2 Small and GPT-3 XL across various
    tasks (i.e., E2E, WebNLG, DART and Curation Corpus) at sparsity levels 50\%
    and 75\% during pre-training. In the metric column, the direction of the
    arrow indicates better result (e.g., up indicates higher is better).}
    \label{tab:alltasks}
    \centering
    \resizebox{\linewidth}{!}{
    \begin{tabular}{cc|ccc|c}
        \toprule
        \multirow{3}{*}{Model} &
        \multirow{3}{*}{\begin{tabular}[c]{@{}c@{}}Pre-Train\\
        Sparsity\end{tabular}} & \multirow{2}{*}{E2E} & \multirow{2}{*}{WebNLG}
        & \multirow{2}{*}{DART} &
        \multirow{2}{*}{\begin{tabular}[c]{@{}c@{}}Curation \\
        Corpus\end{tabular}}  \\ 
        & &                      &                         & & \\ \cmidrule{3-6}
        & & \multicolumn{3}{c|}{BLEU$\uparrow$} & PPL$\downarrow$ \\ \midrule
        \multirow{3}{*}{\begin{tabular}[c]{@{}c@{}}GPT-2 \\ Small\end{tabular}}
        & 0\% & 67.49\textsubscript{$\pm$0.60} & 63.42\textsubscript{$\pm$0.26}
        & 46.30\textsubscript{$\pm$0.16} & 13.38\textsubscript{$\pm$0.02} \\
                              & 50\% &      67.39\textsubscript{$\pm$0.38}     &  
                              63.10\textsubscript{$\pm$0.13}    &
                              45.74\textsubscript{$\pm$0.10}&
                              15.09\textsubscript{$\pm$0.04} \\
                              & 75\% &   66.50\textsubscript{$\pm$0.85}       &
                              62.64\textsubscript{$\pm$0.22}  &
                              44.97\textsubscript{$\pm$0.11} &
                              17.14\textsubscript{$\pm$0.01}            \\
    \midrule \multirow{3}{*}{GPT-3 XL} & 0\% & 68.10\textsubscript{$\pm$0.60} &
        63.62\textsubscript{$\pm$0.23} & 47.71\textsubscript{$\pm$0.11}&
        8.28\textsubscript{$\pm$0.01}   \\
                              & 50\% &    67.98\textsubscript{$\pm$0.63}       &
                              
                              63.47\textsubscript{$\pm$0.21}       &
                              47.10\textsubscript{$\pm$0.13} &
                              9.21\textsubscript{$\pm$0.02}            \\
                              & 75\% &     67.66\textsubscript{$\pm$0.59}      &
                              
                              63.06\textsubscript{$\pm$0.11}        &
                              46.96\textsubscript{$\pm$0.08}&
                              11.03\textsubscript{$\pm$0.02} \\
                              \bottomrule             
    \end{tabular}
    }
\end{table}

\subsection{SPDF on Natural Language Generation and Text Summarization} 

We perform an extended study on SPDF to further investigate its effectiveness on
a diverse set of fine-tuning tasks, when using sparse pre-trained GPT-2 Small
and GPT-3 XL models. In this section, we focus on natural language generation
(i.e., E2E, WebNLG, and DART) and text summarization (i.e., Curation Corpus)
tasks and refer to Table~\ref{tab:alltasks} for all the discussion points. We
note that in Appendix B, we provide evaluation scores on all the metrics used to
officially evaluate E2E, WebNLG and DART, respectively.

First, we validate Hypothesis~\ref{hyp:one} that high degrees of weight sparsity
can be induced during pre-training. Our results indicate that in most settings,
we can pre-train these GPT models with up to 75\% sparsity without significant
degradation across all NLG tasks. On the 75\% sparse GPT-3 XL model, we observe
deltas of -0.44, -0.56, and -0.75 in the BLEU scores for E2E, WebNLG and DART,
respectively. In addition, the 50\% sparse GPT-2 Small model observes deltas of
-0.10, -0.32, and -0.56 in the BLEU scores for E2E, WebNLG and DART,
respectively. Overall, our findings show that these GPT models can be
pre-trained with 50\%-75\% sparsity without losing significant accuracy on these
downstream tasks.

Second, we validate Hypothesis~\ref{hyp:two} that the performance of the sparse
pre-trained model is correlated with the difficulty of the fine-tuning task.
E2E, WebNLG and DART are NLG tasks which focus on mapping structured data
content to a text describing this content. The Curation Corpus task focuses on
summarizing the text description. While both tasks involve generating
semantically coherent natural language, the summarization tasks are more
difficult, since it require understanding of long sequences and compressing the
sequence without loss of information. On the E2E, WebNLG and DART tasks, GPT-3
XL can be pre-trained up to 75\% sparsity without a significant drop in BLEU
score, as discussed previously. In contrast, on Curation Corpus, GPT-3 XL
pre-trained at 75\% sparsity loses 2.75 perplexity. In general, all data-to-text
NLG tasks obtain a lower degradation compared to the more difficult Curation
Corpus summarization task at higher levels of sparsity.

Finally, we validate Hypothesis~\ref{hyp:three} that as the size of the model
increases, it becomes more amenable to higher sparsity levels. We analyze the
relative drop in performance between the dense baseline and its sparse variants
for GPT-2 Small and GPT-3 XL. This trend is clearly evident on the more
difficult Curation Corpus task at 75\% sparsity, where relative to the dense
baseline, the larger GPT-3 XL model has a perplexity delta of +2.75 compared to
a worse +3.76 delta observed in the smaller GPT-2 Small model. Similarly, on the
DART task, the most challenging NLG task out of the three we evaluated, the
delta in the BLEU score is -1.33 and -0.75 for GPT-2 Small and GPT-3 XL,
respectively. These observations indicate that as the size of the language model
increases, it suffers less on downstream task performance when training with
high sparsity.

\subsection{Pre-training vs Fine-tuning Parameter Subspaces}
In this section we analyze the parameter subspaces of the pre-trained model and
its fine-tuned parameters  across all layers to further understand (a) the
behaviour of dense and spare pre-trained representations when fine-tuned, and
(b) the effect of scaling the model size on parameter subspaces between the two
phases. Inspired by~\citet{RadiyaDixit2020HowFC}, we measure the angular
distance (i.e., cosine distance) between the pre-trained model parameters and
its fine-tuned parameters on a given downstream task. Specifically, in all
layers of the language model, we inspect the four weight matrices in the
self-attention module; $W_Q$ (query), $W_K$ (key), $W_V$ (value) and $W_D$
(attention output projection) and the two in the MLP module; $W_I$
(intermediate) and $W_O$ (MLP output projection). In this analysis we focus on
DART, the most difficult NLG task, and report the cosine distances for all
modules in each layer of the dense and 75\% sparse pre-trained GPT-2 Small and
GPT-3 XL. 

\begin{figure}[!t]
    \centering
    \begin{subfigure}[b]{0.78\linewidth}
       \includegraphics[keepaspectratio=true, width=1\linewidth]{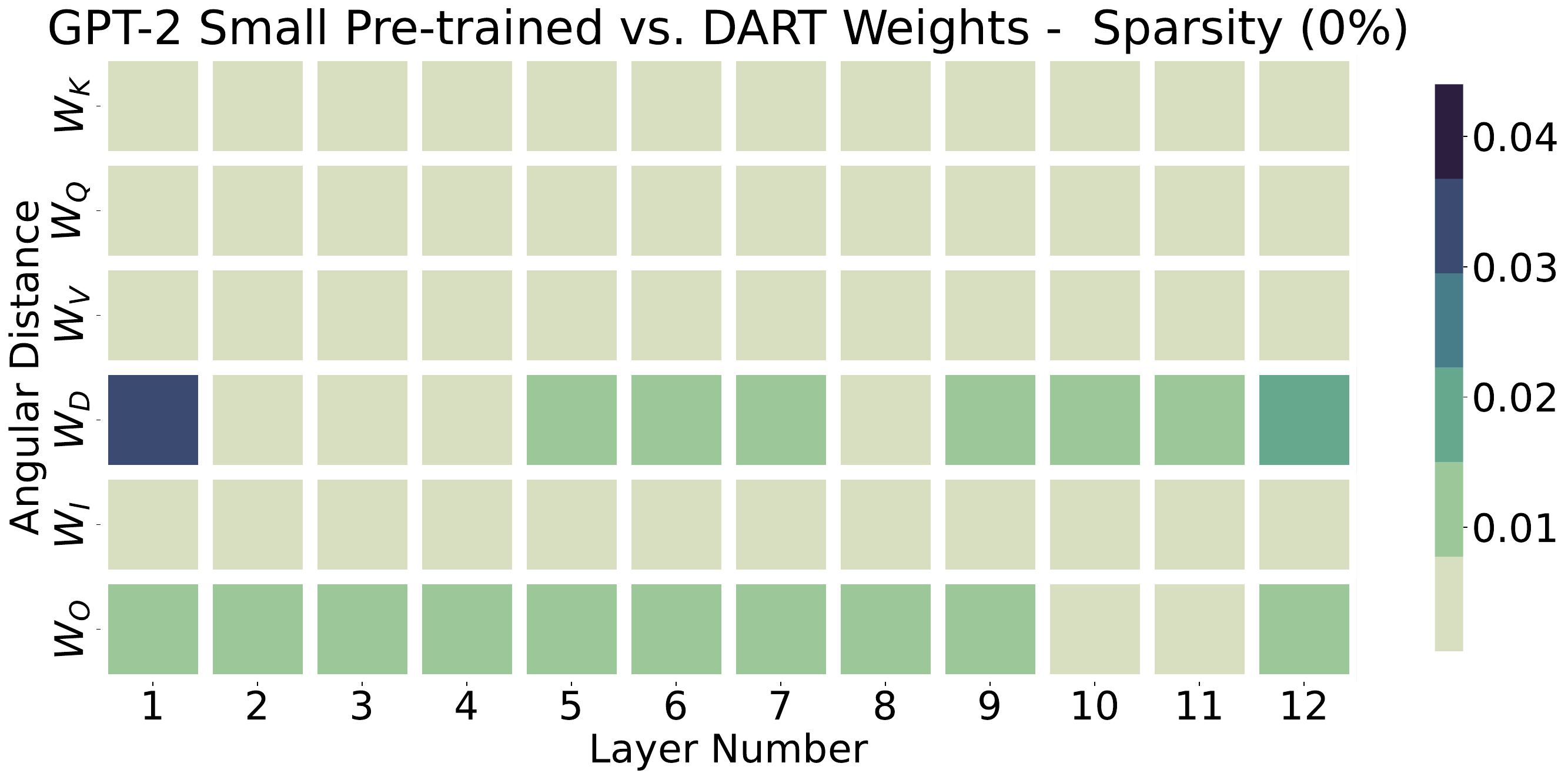}
    \end{subfigure}
    
    \begin{subfigure}[b]{0.78\linewidth}
       \includegraphics[keepaspectratio=true, width=1\linewidth]{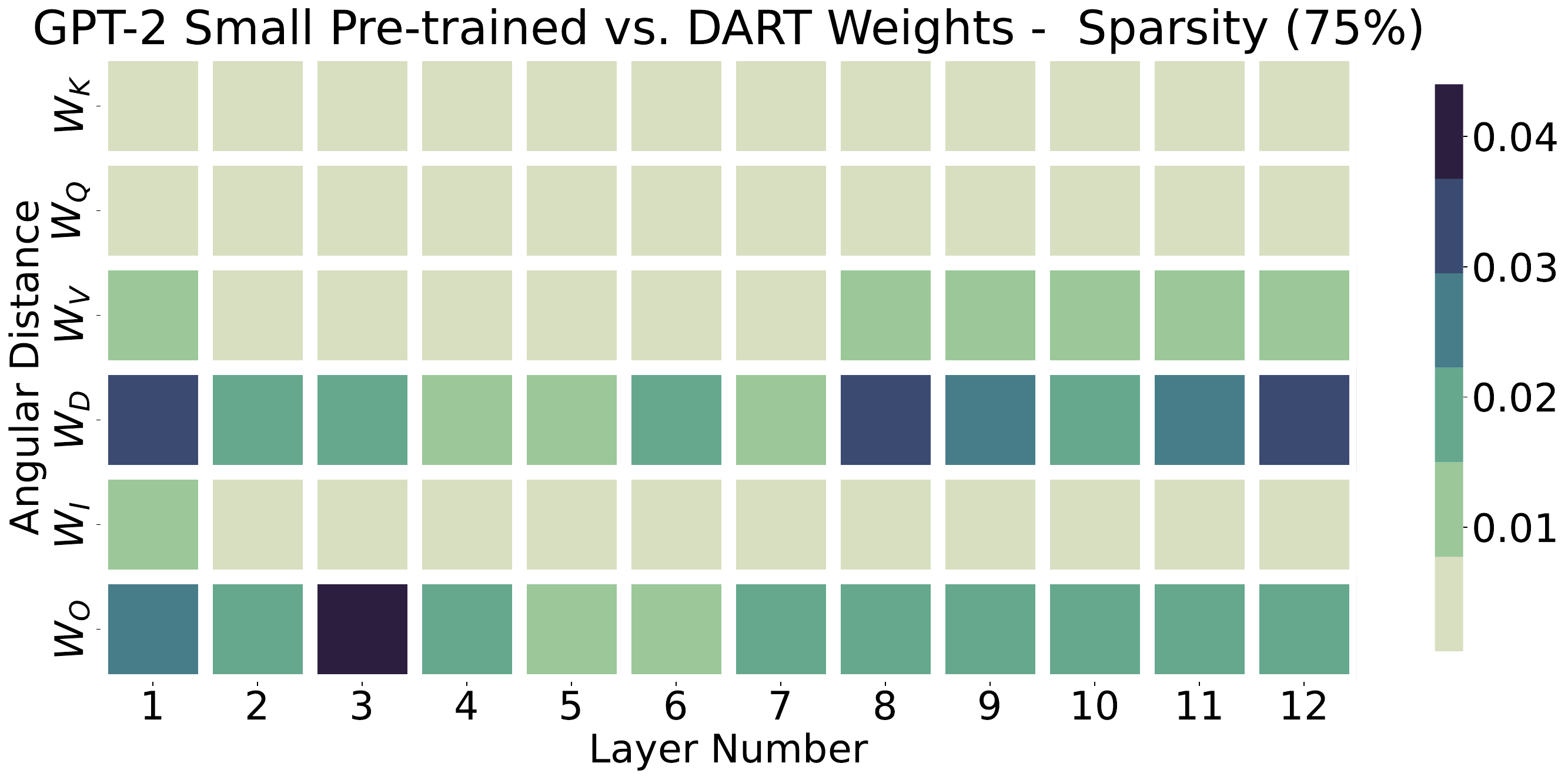}
    \end{subfigure}
    
    \caption[GPT-2 Small Pre-trained vs. DART Weights]{The angular distances in
    parameter subspaces between dense (top) and 75\% sparse (bottom) pre-trained
    and fine-tuned DART weights for GPT-2 Small.}
    \label{fig:gpt2_dart_subspaces} 
\end{figure}

\begin{figure}[!t]
    \centering
    \makebox[\linewidth]{
    \begin{subfigure}[b]{\linewidth}
       \includegraphics[width=1\linewidth]{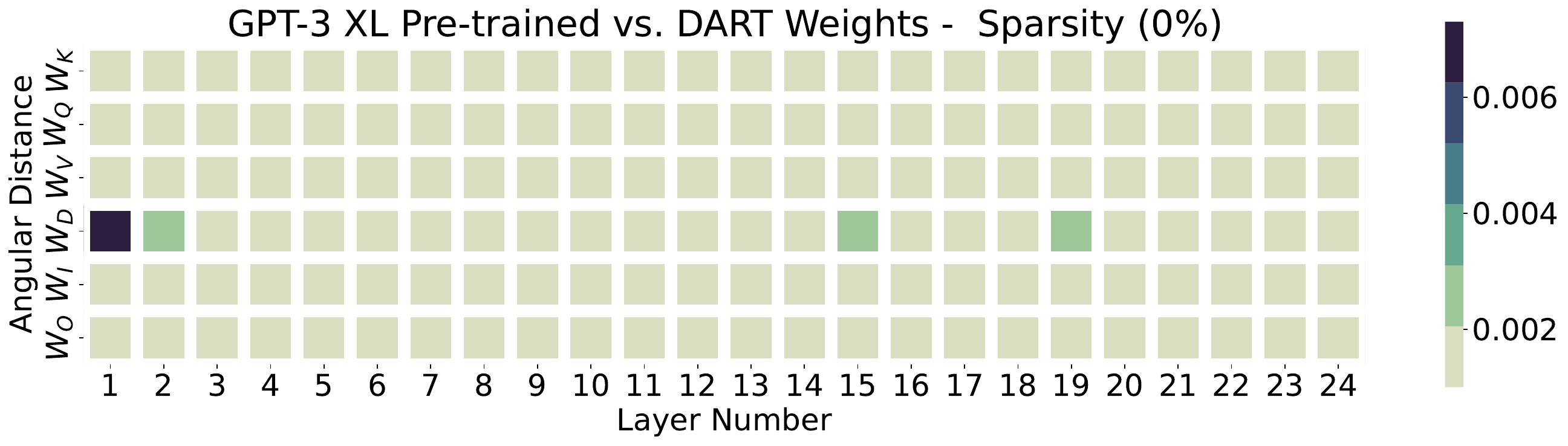}
       
    \end{subfigure}
    }
    \begin{subfigure}[b]{\linewidth}
       \includegraphics[width=1\linewidth]{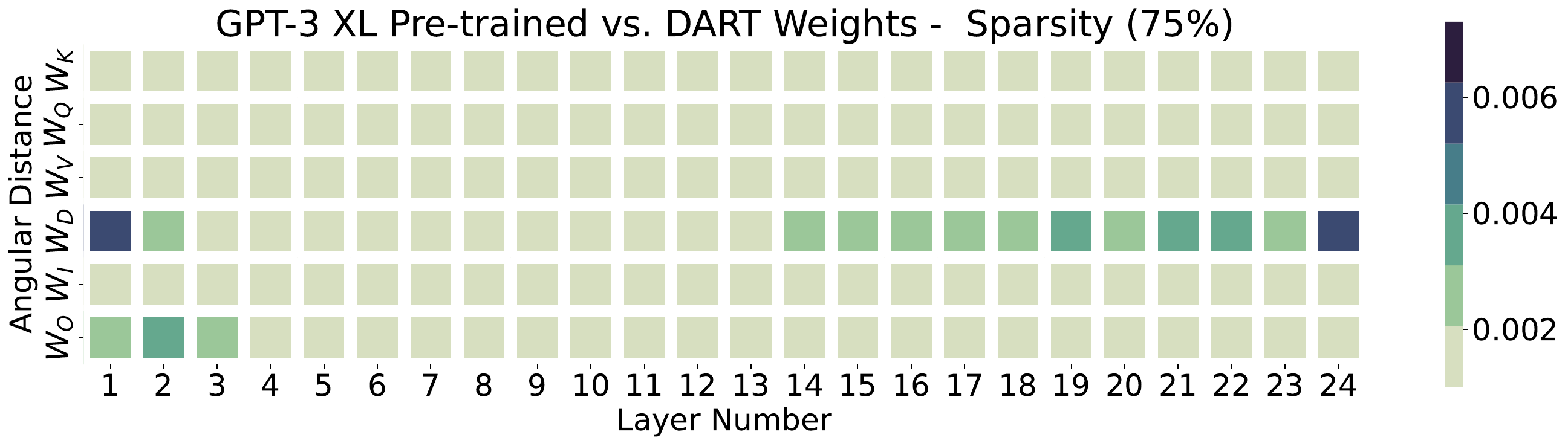}
    \end{subfigure}
    
    \caption[GPT-3 XL Pre-trained vs. DART Weights]{The angular distances in
    parameter subspaces between dense (top) and 75\% sparse (bottom) pre-trained
    and fine-tuned DART weights for GPT-3 XL.}
    \label{fig:gpt3xl_dart_subspaces} 
    \vspace{-0.1in}
\end{figure}

First, we aim to understand the behaviour of the parameter subspaces of the
dense and sparse pre-trained models when fine-tuned. In GPT-2 Small (see
Figure~\ref{fig:gpt2_dart_subspaces}) and GPT-3 XL (see
Figure~\ref{fig:gpt3xl_dart_subspaces}), we observe that the dense pre-trained
parameters and its fine-tuned parameters have very small cosine distances in
almost all modules across each layer, whereas the 75\% sparse model has larger
cosine distances in certain modules (e.g., $W_D$ and $W_O$) across all layers.
Here, the dense model's fine-tuned parameters require less change in the
parameter subspace relative to the pre-trained parameters, while the sparse
model requires more movement in certain modules to learn the downstream task.
This indicates that pre-trained models which learn high quality textual
representations need less movement in the parameter subpsace to adapt to the
downstream task. Although the sparse model has less representational capacity in
its pre-trained parameters, it is capable of adapting certain modules through
dense fine-tuning to learn the downstream task and stay competitive with the
dense model's performance.

Next, we study the effect of model size and the parameter subspaces of the
pre-trained and fine-tuned parameters. Evidently, in
Figure~\ref{fig:gpt3xl_dart_subspaces}, we observe that the dense pre-trained
GPT-3 XL model has very small cosine distances across all modules in almost each
layer, in comparison to GPT-2 Small. This suggests that as we increase the
modeling capacity of the language model, only a few model parameter updates
traverse a very short distance in the parameter space. This results in the
pre-trained and fine-tuned weights being highly close across all modules in
almost each layer. The larger language model is more capable of learning high
quality representations, thus requires less movement in the fine-tuning
parameter subspace. Even at 75\% sparsity, the GPT-3 XL model requires
significantly less change to the pre-trained parameters compared to GPT-2 Small
in order to perform competitively well with the dense model. Given that many
layers experience a very small change in the parameter subspace, we leave the
investigation of freezing these modules during the fine-tuning phase for future
work.

\begin{table*}[!ht]
    \caption{Total FLOPs along with the associated theoretical speedup~w.r.t the
    dense baseline (in brackets) for each of the evaluated fine-tuning tasks on
    GPT-2 Small and GPT-3 XL.  The reported training FLOPs includes both
    pre-training and dense fine-tuning FLOPs. GPT-3 XL 75\% SPDF provides
    $\approx$ 2.5x FLOP reduction over end-to-end dense training.}
    \label{tab:flops_gpt2_gpt3xl}
    \centering \resizebox{0.78\linewidth}{!}{
    \begin{tabular}{cc|cccc}
        \toprule
    \multirow{2}{*}{Model} &
                           \multirow{2}{*}{\begin{tabular}[c]{@{}c@{}}Pre-Train\\
                           Sparsity\end{tabular}} &
                           \multicolumn{4}{c}{Pre-training + Fine-tuning FLOPs
                           ($\times 10^{18}$)}         \\ \cmidrule{3-6} & & E2E
                           & WebNLG       & DART         & Curation Corpus \\
                           \midrule \multirow{3}{*}{GPT-2 Small}           & 0\%
                           & 2.48 (1.00x) & 2.48 (1.00x) & 2.45 (1.00x) & 2.44
                           (1.00x)  \\
                                      & 50\% & 1.84 (1.34x) & 1.82 (1.35x) &
                           1.84 (1.34x) & 1.81 (1.35x) \\
                & 75\% & 1.52 (1.64x)       & 1.49 (1.65x)       & 1.52 (1.64x)
                & 1.48 (1.65x) \\ 
    \midrule \multirow{3}{*}{GPT-3 XL}           & 0\% & 236.62 (1.00x) & 236.62
    (1.00x) & 236.33 (1.00x) & 236.32 (1.00x) \\
               & 50\% & 142.40 (1.66x)       & 142.10 (1.66x) & 142.01 (1.66x) &
               142.40 (1.66x) \\
            & 75\% & 95.29 (2.48x)       & 94.98 (2.49x)        & 95.29 (2.48x)
            & 94.90 (2.49x) \\
\bottomrule      
    \end{tabular}
    }
    \vspace{-0.1in}
\end{table*}

\subsection{SPDF Training Efficiency}
\label{sec:spdf_train_eff}
We compare the standard dense pre-training followed by dense fine-tuning
framework to SPDF and highlight the potential FLOP reduction we can achieve. In
Table~\ref{tab:flops_gpt2_gpt3xl}, we report the total FLOPs (i.e., both the
forward and backward propagations) needed for pre-training and dense fine-tuning
GPT-2 Small and GPT-3 XL models on each of the tasks we evaluated. We note that
in the GPT-2 Small model, the percentage of attention and vocab embeddings FLOPs
account for approximately 13.3\% and 27\% of the total FLOPs, respectively.
Therefore, at 75\%, we achieve approximately 1.65x FLOP reduction over the dense
baseline. However, in the larger GPT-3 XL, the percentage of attention and vocab
embeddings FLOPs account for 13.3\% and 6.8\%, respectively. As a result, at the
GPT-3 XL scale, SPDF provides almost 2.5x FLOP reduction over the dense baseline
when pre-training with 75\% sparsity. The trend of FLOP reduction relative to
the dense baseline continues to increase with larger models, so the potential
gains from sparse pre-training improves as model size grows. We also emphasize
that the total fine-tuning FLOPs is a small fraction of the total pre-training
FLOPs. In Appendix A.4, we provide details on how the total pre-training and
fine-tuning FLOPs for GPT-2 Small and GPT-3 XL were calculated.

\section{Related Work}

\paragraph{Zero-Shot vs. Fine-tuning}
Recent works have shown that large language models can achieve reasonable
performance without any parameter
updates~\citep{brown2020gpt3,chowdhery2022palm,rae2021gopher, smith2022using},
often referred to as the zero-shot or few-shot setting. When no parameters are
fine-tuned, framing a target task in terms of the pre-training objective enables
zero-shot or few-shot learning to use a task-specific prompt and a few examples
of a task~\citep{brown2020gpt3}. However, while such few-shot learning is simple
using such large models, there are alternative methods to obtain similar task
accuracy using smaller models~\citep{schick2021smallllm}. In recent
work,~\citet{lamda2022} demonstrate that while scaling the size of LaMDA can
improve quality, combining scaling with fine-tuning can improve the model across
all metrics including quality, safety and groundness.~\citet{irene2021} show
that fine-tuning also helps update language model behaviour to mitigate harmful
outputs, which is highly critical for real-world deployment of LLMs (e.g.,
ChatGPT~\citep{openai_chatgpt_2022}, Bard~\citep{pichai_2023}, etc.). To achieve
the best performance in practice, fine-tuning will continue to be the modus
operandi when using pre-trained LLMs. Hence, our work focuses on pre-training
and fine-tuning language models across a diverse set of tasks, including natural
language generation and text summarization.

\paragraph*{Efficient Fine-tuning} While most large-scale models such as
GPT~\citep{brown2020gpt3, smith2022using} or T5~\citep{raffel2022transfer} are
trained dense, there are works~\citep{houlsby2019parameter, li2021prefix,
zaken2021bitfit, hu2022lora} that explore using limited capacity (tuning a few
layers or subset of parameters) in the pre-trained models to fine-tune on
downstream tasks. These works are indicative that the total modeling capacity is
unnecessary for fine-tuning on downstream tasks. Our work draws some inspiration
from these works for exploiting the limited capacity of models for final tasks.
However, we choose to reduce FLOPs for pre-training (significantly more training
FLOPs than fine-tuning) and then add all the modeling capacity back during
fine-tuning. This allows us to train large models efficiently and yet retain
accuracies comparable to dense baselines. Although we do not explore efficient
fine-tuning in our study, we leave the exploration of using alternative sparsity
schedules~\citep{zhu2018toprune, liu2021sparsegranet}, adapting a subset of
parameters during fine-tuning~\citep{Ding2022DeltaTA} and imposing low-rank
structures~\citep{hu2022lora} for future work.

\paragraph{Weight Sparsification Techniques}
Many unstructured weight sparsification techniques have been proposed in the
literature for training neural networks~\citep{hoefler2022sparsity}, which can
be categorized as static sparsity and dynamic sparsity. Static sparsity methods
have a fixed sparsity structure (i.e., sparsity mask) determined at
initialization~\citep{lee2018snip,Wang2020Picking}. In contrast, dynamic sparse
training (DST) methods iteratively prune (drop) and add (regrow) weights during
training~\citep{mocanu2018, evci2020rigl, jayakumar2020top,
shaoyi2022betterrigl} to find the best possible sparse subnetwork while
retaining accuracy comparable to dense baselines. Although, dynamic sparse
training methods can help achieve Pareto improvements in terms of number of
training FLOPs to accuracy, we leave this for future work. Inspired
by~\citep{li2022the}, which shows that scaling the size of CNNs closes the gap
between a randomly pruned sparse network and its dense counterpart, we focus our
study on language models with static sparsity. 
While~\citet{dao2022monarch} demonstrate the benefits of sparse-to-dense
training, they mainly apply it during pre-training and instead, focus their
studies on dense-to-sparse fine-tuning similar to other efficient fine-tuning
efforts. In our work, we show that sparse pre-training followed by dense
fine-tuning on downstream tasks can be on par with the accuracy of a dense
pre-trained model on many tasks, while significantly lowering overall training
FLOPS.

\section{Conclusion and Future Work}

In this work, we introduced Sparse Pre-training and Dense Fine-tuning (SPDF) to
reduce the computational FLOPs of training GPT models using weight sparsity. To
the best of our knowledge, this is the first time a large GPT model has been
pre-trained with high sparsity (50\%-75\%) without significant loss in
downstream task metrics. In our work, we only use simple static sparsity, which
is arguably the most naïve way to induce sparsity in neural networks. As for
future work, there are several natural directions for improving our results on
even larger models, including dynamic sparsity methods, better optimization
techniques for sparse training, and architectures amenable to sparse training.
Moreover, to limit the computational cost of our study, we trained our GPT
models following the Chinchilla scaling law. Although the Chinchilla
pre-training schedule has been shown to be FLOP-optimal for dense models, we
plan to investigate how well it transfers to sparse models. Our future work will
also investigate sparse scaling outside the Chinchilla dense scaling laws.
Regardless, we see the tremendous promise of unstructured weight sparsity to
accelerate the training of LLMs, enabled by the recent advances in deep learning
hardware accelerators.

\begin{contributions} 
We provide a summary of each author’s contributions:
\begin{itemize}
    \item Vithursan Thangarasa led the effort for training/evaluation of large scale GPT
    models on the Cerebras CS-2, evaluated the technique in different FLOP efficient
    training setups, brought up multiple downstream tasks, analyzed the parameter
    subspaces, and wrote the manuscript.
    
    \item Abhay Gupta helped with pre-training GPT models on the CS-2 and ran reference
    models to validate our training and fine-tuning setup.
    
    \item William Marshall brought up various downstream tasks on the CS-2 and assisted in
    running fine-tuning experiments.
    
    \item Tianda Li assisted William Marshall and Vithursan Thangarasa with running
    fine-tuning experiments.
    
    \item Kevin Leong assisted Abhay Gupta with pre-training GPT models on the CS-2 and
    provided crucial help in debugging issues.
    
    \item Dennis DeCoste conceived the original key idea.
    
    \item Sean Lie coordinated the bring up of GPT on CS-2 and was involved in
    experimental validation and analysis.
    
    \item Shreyas Saxena advised the entire effort, brought up the initial proof of
    concept and experimented with different sparsity techniques.
    
    \item Shreyas Saxena and Sean Lie frequently met with Vithursan Thangarasa to discuss
    the work and helped revise several iterations of the manuscript.
\end{itemize}

\end{contributions}

\section{Acknowledgements}
We thank Anshul Samar, Dimitrios Sinodinos, and Joel Hestness, for helpful edits
and suggestions that improved the clarity of our manuscript.

\clearpage
\bibliography{refs}

\clearpage


\section*{Supplementary material}

# SPDF: Sparse Pre-training and Dense Fine-tuning for Large Language Models (Supplementary Material)

Vithursan Thangarasa[1]    Abhay Gupta[1]    William Marshall[1]    Tianda Li[*]    Kevin Leong[1]

Dennis DeCoste[*]    Sean Lie[1]    Shreyas Saxena[1]

[1]Cerebras Systems Inc., Sunnyvale, California, USA

# APPENDIX

## A  EXPERIMENTAL SETUP AND HYPERPARAMETER DETAILS

### A.1  PRE-TRAINING ON PILE

To train all GPT models, we use AdamW optimizer [Loshchilov and Hutter, 2017] with $\beta_1 = 0.9$, $\beta_2 = 0.999$ and $\epsilon = 10^{-8}$. The global norm is clipped at 1.0, and a weight decay of 0.1 is used. There is a learning rate warm-up over the first 375M tokens, followed by a cosine decay to 10% of the peak learning rate. In Table 1, we provide details on the size and architecture configurations of the models we pre-trained. Here, $n_{params}$ is the total number of trainable parameters, $n_{layers}$ is the number of decoder layers, and $d_{model}$ is the base size of the model. The feedforward bottleneck is four times the base size, i.e., $d_{\text{ff}} = 4 \times d_{model}$. Finally, $n_{heads}$ are the number of attention heads and $d_{head}$ is the dimension of each attention head. The context window size is set to 2048 following [Brown et al., 2020].

Table 1: Sizes, architectures, and learning hyperparameters (batch size and learning rate) of the models, which are trained to Chinchilla optimal configurations ($\approx$ 20 tokens per parameter).

| Model | $n_{params}$ | $n_{layers}$ | $d_{model}$ | $n_{heads}$ | $d_{head}$ | Batch Size | Learning Rate | Training Tokens |
|---|---|---|---|---|---|---|---|---|
| GPT-2 Small | 125M | 12 | 768 | 12 | 64 | 256 | 6e-4 | 2.5B |
| GPT-3 XL | 1.3B | 24 | 2048 | 16 | 128 | 512 | 2e-4 | 26B |

**Sparsity Setup**   As mentioned in Section 2, we use random pruning (static sparsity) for training all our GPT models. The sparsity is distributed uniformly across all layers (i.e., all layers get the same specified sparsity level) irrespective of the number of parameters or FLOPs of a given layer. We only sparsify all dense linear layers including the two in the MLP module; $W_I$ (intermediate) and $W_O$ (MLP output projection), and the four weight matrices in the self-attention module; $W_Q$ (query), $W_K$ (key), $W_V$ (value) and $W_D$ (attention output projection). In this setting, we keep the embeddings (implemented as sparse lookups), LayerNorm [Ba et al., 2016], and biases dense.

### A.2  NATURAL LANGUAGE GENERATION

Similar to Hu et al. [2022], we train all of our GPT-2 Small and GPT-3 XL models using AdamW [Loshchilov and Hutter, 2017] with a linear learning rate schedule for 5 epochs, and perform early-stopping when the models began to overfit. We perform a grid search to discover an appropriate learning rate that led to the best downstream BLEU score on each of the tasks for a given compute budget. More specifically, on the dense baseline and sparse variants, we select the best batch size among {8, 16, 32, 64} and select the best learning rate among {1e-4, 5e-5, 2.5e-5} on the validation set. The beam search beam size, length penalty, and no repeat ngram size remain the same as described in [Hu et al., 2022].

Table 2: Details on the total pre-training FLOPs for each pre-trained model.. We also report the relative FLOPs reduction of the sparse variants over the dense baseline. We note that the reported FLOPs per sequence (seq) includes both forward and backward passes.

| Model | Pre-Train Sparsity | Total Seqs | Total FLOPs/ Seq | Total FLOPs | Toal exaFLOPs | FLOPs Reduction over Dense |
|---|---|---|---|---|---|---|
| GPT-2 Small | 0% | 1.22e6 | 1.99e12 | 2.43e18 | 2.43 | 1x |
| | 50% | 1.22e6 | 1.47e12 | 1.79e18 | 1.79 | 0.737x |
| | 75% | 1.22e6 | 1.20e12 | 1.46e18 | 1.46 | 0.601x |
| GPT-3 XL | 0% | 1.27e7 | 1.86e13 | 2.36e20 | 236.10 | 1x |
| | 50% | 1.27e7 | 1.12e13 | 1.42e20 | 141.87 | 0.601x |
| | 75% | 1.27e7 | 7.46e12 | 9.48e19 | 94.76 | 0.401x |

### A.3 CURATION CORPUS

We fine-tune the dense GPT-2 Small and its sparse variants on the Curation Corpus [Curation, 2020] following the setup presented in [Rae et al., 2021]. We fine-tune for 5 epochs on Curation Corpus, and perform early stopping once the models start to overfit. To discover good hyperparameters, we perform a grid search to discover an appropriate learning rate that led to the best perplexity for a given compute budget. More specifically, on the dense baseline and sparse variants, we use a batch size of 32 and select the best learning rate among {3e-3, 1e-3, 3e-4, 1e-4, 3e-5, 1e-5} on the validation set.

### A.4 TRAINING FLOPS FOR PRE-TRAINING AND FINE-TUNING

We compute the total pre-training FLOPs for the dense and sparse variants of GPT-2 Small and GPT-3 XL, and report them in Table 2, along with their relative FLOPs reduction over the dense baseline. We also compute the total dense fine-tuning FLOPs for GPT-2 Small and GPT-3 XL on E2E, WebNLG, DART and Curation Corpus, and report them in Table 3. The total training FLOPs during the fine-tuning phase is a small fraction of the total pre-training FLOPs, even though fine-tuning is performed in a dense manner.

Table 3: Details on the total fine-tuning FLOPs for GPT-2 Small and GPT-3 XL on E2E, WebNLG, DART and Curation Corpus tasks. We note that the reported FLOPs per sequence (seq) includes both forward and backward passes.

| Dataset | Model | Total Seq | Total FLOPs/ Seq | Total FLOPs | Toal exaFLOPs |
|---|---|---|---|---|---|
| E2E | GPT-2 Small | 1.26e5 | 1.36e11 | 5.15e16 | 0.052 |
| | GPT-3 XL | | 1.39e12 | 5.27e17 | 0.524 |
| WebNLG | GPT-2 Small | 0.54e5 | 1.36e11 | 2.21e16 | 0.022 |
| | GPT-3 XL | | 1.39e12 | 2.26e17 | 0.226 |
| DART | GPT-2 Small | 1.25e5 | 1.36e11 | 5.12e16 | 0.051 |
| | GPT-3 XL | | 1.39e12 | 5.24e17 | 0.524 |
| Curation Corpus | GPT-2 Small | 0.34e5 | 1.36e11 | 1.38e16 | 0.014 |
| | GPT-3 XL | | 1.39e12 | 1.41e17 | 0.141 |

## B   DETAILED EXPERIMENTS ON NLG AND TEXT SUMMARIZATION

In Tables 4, 5 and 6, we provide detailed results on all official evaluation metrics for the E2E, WebNLG and DART tasks. Even across other metrics, we do not observe any significant drop in performance with sparse pre-trained GPT-2 Small and GPT-3 XL models.

Table 4: Downstream accuracy of GPT-2 Small and GPT-3 XL on E2E at different sparsity levels during pre-training. In the metric column, the direction of the arrow indicates better result (e.g., up indicates higher is better).

| Model | Pre-Train Sparsity | E2E | | | | |
|---|---|---|---|---|---|---|
| | | BLEU↑ | NIST↑ | METEOR↑ | ROUGE-L↑ | CIDEr↑ |
| GPT-2 Small | 0% | $67.49_{\pm 0.60}$ | $8.59_{\pm 0.03}$ | $46.08_{\pm 0.22}$ | $70.22_{\pm 0.42}$ | $2.38_{\pm 0.03}$ |
| | 50% | $67.39_{\pm 0.38}$ | $8.62_{\pm 0.03}$ | $45.89_{\pm 0.18}$ | $70.10_{\pm 0.26}$ | $2.38_{\pm 0.01}$ |
| | 75% | $66.50_{\pm 0.01}$ | $8.46_{\pm 0.13}$ | $45.61_{\pm 0.32}$ | $70.02_{\pm 0.26}$ | $2.36_{\pm 0.02}$ |
| GPT-3 XL | 0% | $68.10_{\pm 0.46}$ | $8.64_{\pm 0.06}$ | $46.40_{\pm 0.01}$ | $71.03_{\pm 0.22}$ | $2.41_{\pm 0.02}$ |
| | 50% | $67.98_{\pm 0.63}$ | $8.62_{\pm 0.07}$ | $46.40_{\pm 0.07}$ | $71.30_{\pm 0.20}$ | $2.43_{\pm 0.02}$ |
| | 75% | $67.66_{\pm 0.59}$ | $8.59_{\pm 0.09}$ | $46.07_{\pm 0.26}$ | $70.40_{\pm 0.26}$ | $2.42_{\pm 0.03}$ |

Table 5: Downstream BLEU, NIST, METEOR, ROUGE-L and CIDEr scores of GPT-2 Small and GPT-3 XL on WebNLG at different sparsity levels during pre-training. In the metric column, the direction of the arrow indicates better result (e.g., down indicates lower is better).

| Model | Pre-Train Sparsity | WebNLG | | |
|---|---|---|---|---|
| | | BLEU↑ | METEOR↑ | TER↓ |
| GPT-2 Small | 0% | $63.42_{\pm 0.26}$ | 0.44 | 0.34 |
| | 50% | $63.10_{\pm 0.13}$ | 0.44 | 0.34 |
| | 75% | $62.64_{\pm 0.22}$ | 0.43 | 0.34 |
| GPT-3 XL | 0% | $63.62_{\pm 0.23}$ | 0.45 | 0.32 |
| | 50% | $63.47_{\pm 0.21}$ | 0.45 | 0.33 |
| | 75% | $63.06_{\pm 0.11}$ | 0.45 | 0.33 |

Table 6: Downstream BLEU, MET and TER scores of GPT-2 Small and GPT-3 XL on DART at different sparsity levels during pre-training. In the metric column, the direction of the arrow indicates better result (e.g., down indicates lower is better).

| Model | Pre-Train Sparsity | DART | | |
|---|---|---|---|---|
| | | BLEU↑ | MET↑ | TER↓ |
| GPT-2 Small | 0% | $46.30_{\pm 0.16}$ | 0.38 | 0.51 |
| | 50% | $45.74_{\pm 0.10}$ | 0.37 | 0.51 |
| | 75% | $44.97_{\pm 0.11}$ | 0.37 | 0.52 |
| GPT-3 XL | 0% | $47.71_{\pm 0.11}$ | 0.39 | 0.49 |
| | 50% | $47.10_{\pm 0.13}$ | 0.39 | 0.49 |
| | 75% | $46.96_{\pm 0.08}$ | 0.38 | 0.50 |

## C  UNSTRUCTURED SPARSITY ON SPECIALIZED HARDWARE ACCELERATORS

In Figure 1 we highlight the potential realized gains with unstructured weight sparsity on the Cerebras CS-2. This figure was regenerated based on the plot in [Lie, 2021].

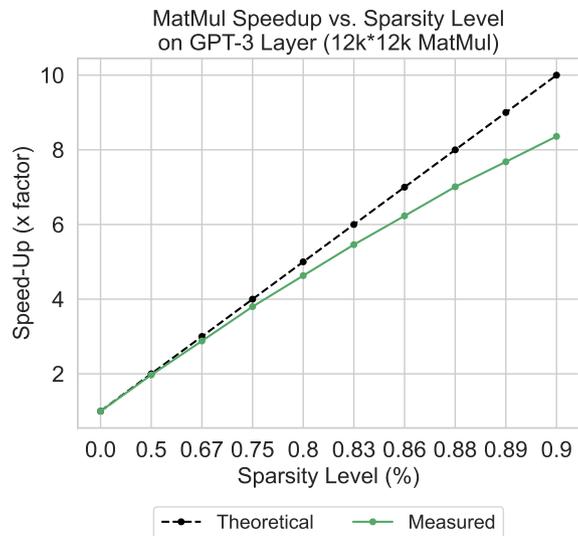

Figure 1: Measured speedup versus theoretical speedup at varying sparsity levels for a GPT-3 layer 12k × 12k matrix multiplication (MatMul) [Lie, 2021].



\end{document}


\onecolumn 
\maketitle

This Supplementary Material should be submitted as a separate file. Please do not append the Supplementary Material to the main paper. 

Fig. \ref{fig:pitt} and Eq \ref{eq:example} in the main paper can be cross referenced using \texttt{xr}. 

\appendix
\section{Additional simulation results}
Table~\ref{tab:supp-data} lists additional simulation results; see also \citet{einstein} for a comparison. 

\begin{table}[!h]
    \centering
    \caption{An Interesting Table.} \label{tab:supp-data}
    \begin{tabular}{rl}
      \toprule 
      \bfseries Dataset & \bfseries Result\\
      \midrule 
      Data1 & 0.12345\\
      Data2 & 0.67890\\
      Data3 & 0.54321\\
      Data4 & 0.09876\\
      \bottomrule 
    \end{tabular}
\end{table}

\section{Math font exposition}
\providecommand{\upGamma}{\Gamma}
\providecommand{\uppi}{\pi}
How math looks in equations is important:
\begin{equation*}
  F_{\alpha,\beta}^\eta(z) = \upGamma(\tfrac{3}{2}) \prod_{\ell=1}^\infty\eta \frac{z^\ell}{\ell} + \frac{1}{2\uppi}\int_{-\infty}^z\alpha \sum_{k=1}^\infty x^{\beta k}\mathrm{d}x.
\end{equation*}
However, one should not ignore how well math mixes with text:
The frobble function \(f\) transforms zabbies \(z\) into yannies \(y\).
It is a polynomial \(f(z)=\alpha z + \beta z^2\), where \(-n<\alpha<\beta/n\leq\gamma\), with \(\gamma\) a positive real number.

\bibliography{uai2023-template}